\documentclass[11pt]{article}
\pdfoutput=1
\usepackage{enumitem}
\usepackage[utf8]{inputenc}
\usepackage[english]{babel}
\usepackage[a4paper]{geometry}

\usepackage[normalem]{ulem}

\usepackage{amssymb}
\usepackage{mathtools,xparse}
\usepackage[lined,boxed]{algorithm2e}

\usepackage{tcolorbox}

\usepackage{graphicx}

\usepackage{lipsum}
\usepackage{float}
\usepackage{subcaption}
\graphicspath{ {./images/} }
\usepackage[export]{adjustbox}

\numberwithin{equation}{section}
\usepackage{hyperref}
\usepackage{booktabs}
\usepackage{tabularx}

\newcommand{\FrameboxA}[2][]{#2}
\newcommand{\Framebox}[1][]{\FrameboxA}

\newcommand{\bfA}{{\bf A}}

\newcommand{\bfD}{{\bf D}}

\newcommand{\bfG}{{\bf G}}
\newcommand{\bfH}{{\bf H}}

\newcommand{\bfJ}{{\bf J}}

\newcommand{\bfP}{{\bf P}}

\newcommand{\bfX}{{\bf X}}

\newcommand{\bfe}{{\bf e}}
\newcommand{\bfb}{{\bf b}}

\newcommand{\bfx}{ {\bf x}}

\newcommand{\bfz}{ {\bf z}}

\newcommand{\bfu}{{\bf u}}

\newcommand{\bfd}{{\bf d}}
\newcommand{\bfm}{{\bf m}}
\newcommand{\bfr}{{\bf r}}

\newcommand{\bfg}{{\bf g}}

\definecolor{darkblue}{rgb}{0.08, 0.15, 0.48}

\usepackage[titles]{tocloft}
\usepackage[multiple]{footmisc}

\newcommand\blfootnote[1]{%
  \begingroup
\renewcommand\thefootnote{}\footnote{#1}%
  \addtocounter{footnote}{0}%
  \endgroup
}

\title{
\textbf{Physics-guided Full Waveform Inversion using Encoder-Solver Convolutional Neural Networks}}
\author{\blfootnote{The Israel Science Foundation supported this research grant no. 656/23 and by the
Lynn and William Frankel Center for Computer Science at BGU. MG is supported by the Kreitman
High-Tech scholarship.}
Matan M. Goren, Eran Treister
\blfootnote{Computer Science Department, Ben-Gurion University of the Negev, Beer Sheva, Israel.\\
(gorenm@post.bgu.ac.il, erant@cs.bgu.ac.il)}\thefootnote{}}

\date{}
\begin{document}  
\maketitle
\begin{abstract}
Full Waveform Inversion (FWI) is an inverse problem for estimating the wave velocity distribution in a given domain, based on observed data on the boundaries. 
The inversion is computationally demanding because we are required to solve multiple forward problems, either in time or frequency domains, to simulate data that are then iteratively fitted to the observed data. We consider FWI in the frequency domain, where the Helmholtz equation is used as a forward model, and its repeated solution is the main computational bottleneck of the inversion process. To ease this cost, we integrate a learning process of an encoder-solver preconditioner that is based on convolutional neural networks (CNNs). The encoder-solver is trained to effectively precondition the discretized Helmholtz operator given velocity medium parameters. Then, by re-training the CNN between the iterations of the optimization process, the encoder-solver is adapted to the iteratively evolving velocity medium as part of the inversion. Without retraining, the performance of the solver deteriorates as the medium changes. Using our light retraining procedures, we obtain the forward simulations effectively throughout the process. We demonstrate our approach to solving FWI problems using 2D geophysical models with high-frequency data.
\end{abstract}

\section{Introduction}

Full waveform inversion (FWI) is an inverse problem where we estimate the wave propagation velocity based on recorded seismic data on the boundaries. The solution of FWI was originally developed for geophysics purposes such as sub-surface mapping and seismic exploration of oil and gas reservoirs, but was shown to be applicable for other imaging tasks including brain imaging \cite{guasch2020brain}, ultrasound imaging \cite{bernard2017ultrasonictomography} and optical diffraction tomography \cite{soubies2017diffraction}. The problem is formulated as an optimization problem and is solved by a descent algorithm which iteratively updates the underlying velocity model to reduce the misfit function, measuring the difference between the recorded seismic data and the simulated
waveforms. At each iteration, the waveform simulation is obtained for multiple sources for a given velocity model. This is done by either time integrating through the wave equation (in the time domain) or solving the Helmholtz equation (in the frequency domain) for multiple frequencies. That is, in the frequency domain, which is the focus of this work, one needs to solve the Helmholtz equation for multiple right-hand sides (sources) and multiple frequencies at each iteration. 

The efficient numerical solution of the Helmholtz equation with high and spatially-dependent wavenumber is a difficult computational task. The linear system obtained from the discretization of the Helmholtz equation involves a very large, complex-valued, sparse, and indefinite matrix. The equation is most difficult in three dimensions, where a direct solver is highly memory-consuming and impractical in most scenarios. On the other hand, there are multiple available iterative methods to solve the equation, but they typically require many iterations at high wavenumbers. Either way, the data simulations are highly cumbersome in FWI and are the main computational bottleneck of the inversion. 

Beyond the high computational costs of the forward simulations, FWI is a highly non-linear and ill-posed problem, especially in the absence of low-frequency data, which are hard to acquire because of physical limitations \cite{demanet2016lowfrequency}. Therefore, sophisticated regularization techniques \cite{eran2021extendedfwi}, loss functions \cite{johnson2016perceptualloss}, and careful minimization strategies \cite{treister2017full,bao2023robust} are often considered. Still, even after many years of research, the solution process of FWI is sensitive to the inversion hyper-parameters, and cumbersome computationally. 

In recent years, data-driven deep learning (DL) methods \cite{krizhevsky2017imagenet,lecun2015deep,szegedy2015going} have revolutionized many scientific fields, most notably computer vision and natural language processing. Inverse problems were also treated by DL, and FWI in particular, because of their difficulty and non-linearity. In the context of FWI, two main approaches are considered: (1) data-guided approaches (end-to-end) where deep networks are trained to consume data on the one hand and output a velocity model on the other \cite{wulin2019inversionnet,feng2023simplifying,Li2020}, and (2) physics-guided approaches where the data-fitting structure of the problem is kept and we make sure that the resulting model fits the observed data in simulations \cite{sanghvi2019embedding,colombo2023self}. Here, we focus on the physics-guided approach, since the first approach is highly dependent on the setup of the problem used during training, which may be incompatible with real-life scenarios. This includes locations of sources and receivers, types of media in the training set, missing data due to technical problems, etc. In other words, end-to-end training may not be flexible enough to handle different experiments. In the physics-guided approach, we can be flexible with respect to the setup of the problem, and we are faithful to the classical approach of fitting numerical simulations. The latter is important for applications where the confidence and correctness of the velocity estimation are more important than the computational cost of the inversion, like in oil and gas exploration. In such applications, in particular, the estimation is not required in near-real-time, like in other vision applications that exploit the efficiency of the end-to-end approach. Choosing the physics-guided approach, we integrate deep learning within the optimization process of solving FWI.

In this work, we tackle the computational cost of the physical simulations in FWI, by using the multigrid-augmented encoder-solver convolutional neural network (CNN) approach to solve the Helmholtz equation, where two convolutional neural networks are used to solve the equation \cite{azulay2022multigrid,lerer2024multigridaugmented}. The encoder network is used once-per-medium to encode context vectors, and the solver network is used as a preconditioner to solve the system for the many right-hand-sides, given the context vectors. The main observation in \cite{azulay2022multigrid} is that for a single given medium, the solver CNN can solve the linear system efficiently to any accuracy, but it struggles when it is expected to generalize over the velocity medium, which is generally unknown during training. Hence, the advantage of the approach in our case stems from two properties of the inverse solution:
\begin{enumerate}
\item At each iteration of the inverse solution, we solve multiple systems with the same medium but with different right-hand-sides. That is true even if we use acceleration techniques such as simultaneous sources \cite{haber2012effective}, and even more so if we use high order methods like Gauss-Newton \cite{pratt1998gauss}.
\item During the inversion, the velocity model changes gradually, hence the network's weights do not need to change dramatically to cope with that and maintain the solver's efficiency. Hence, small training sessions between the inversion iterations can dramatically speed up the solution process, relieving the need for the CNN to generalize over velocity mediums.
\end{enumerate}

To summarize, in this work we use the encoder-solver deep learning approach to accelerate physics-guided FWI. We accelerate the simulations to ease the overall inversion process significantly (e.g., inversion parameter tuning). We do note that DL approaches can be highly beneficial in other aspects of FWI as well, like regularization and data pre-processing \cite{demanet2016lowfrequency}, in addition to our approach.

The rest of the work is organized as follows. In the following sections, we review related work and discuss preliminaries and background that are needed to introduce our approach. Next, we present our main contribution: a light-weight data efficient retraining procedure of the encoder-solver model. Lastly, we demonstrate the improvements and effectiveness of our approach when solving the FWI problem. We include the results of experiments performed on several 2D models with high-frequency data.

\section{Related Work}
\paragraph{End-to-end neural network FWI solvers.} With the advances of DL methods for the solution of inverse problems \cite{lucas2018using,liang2020deep}, and the concurrent rise in recorded seismic data, efficient DL models have been proposed to solve the FWI problem. The DL approach involves training a neural network and the solution is obtained by the network's forward application. That is, the DL model is viewed as an end-to-end solver and entirely replaces the classical inversion process. Various models have been proposed, including Multi-Layer Perceptron (MLP) \cite{sun2023implicit,kim2018geophysical}, Convolutional Neural Network (CNN) \cite{araya2018deep,park2020automatic,wulin2019inversionnet,jin2021unsupervised}, Recurrent Neural Network (RNN) \cite{richardson2018seismic,adler2019deep} and encoder-decoder models \cite{gelboim2022encoder,gelboim2023deep,feng2023simplifying, Li2020}. However, end-to-end data driven methods are highly dependent on the data and the training process of the network. As a result, this approach suffers from a generalization gap when faced with out-of-distribution data. Moreover, acquiring rich and expressive training data is a difficult task that impacts the generalization capabilities of the proposed models.

\paragraph{Physics-informed neural network.} To overcome the challenges of training the neural network posed by the data acquisition process, \cite{raissi2019physics} proposed a physics-informed neural network (PINN) instead of using a pure data-mapping loss objective. In PINN, the underlying physical laws is incorporated in the loss function to reconstruct NN-based functional solutions to PDEs. That is, given spatial or temporal input data, the network estimates the solution to the PDE at the coordinates of interest. PINNs use automatic differentiation
to represent all the differential operators \cite{baydin2018automatic} and hence there is
no explicit need for a mesh or grid data representation \cite{bar2019unsupervised,torres2022mesh}. In geophysical applications, among others, the PINN approach was shown to be effective for solving the forward problem, given by the Eikonal equation \cite{waheed2020eikonal} or by the Helmholtz equation \cite{alkhalifah2020wavefield,song2021solving,song2022versatile,stanziola2021helmholtz}, and for solving the inversion problem \cite{sun2023implicit,muller2023deep,rasht2022physics}.

\paragraph{Classical inversion enhanced by deep learning.} A different approach aims to improve the performance of classical methods by integrating DL networks in the inversion process. One of the proposed improvements is to mitigate the issue of cycle-skipping in FWI \cite{virieux2009overview}. \cite{sun2020extrapolated,ovcharenko2019deep} proposed a CNN-based model for low-frequency data extrapolation. \cite{bao2023robust} proposed to perform wavelet source transformation \cite{wang2021seismic} using a CNN model, which results in a more robust inversion process. A different approach was proposed by \cite{richardson2018seismic}, where an RNN is utilized to model the forward wave equation and the optimization is solved using stochastic approaches, such as ADAM \cite{kingma2014adam} or SGD. However, the inversion process is still computationally demanding and many iterations are required to obtain a good velocity model estimation. Here, we propose to improve the forward simulation, obtained by the solution of the Helmholtz equation, using the encoder-solver architecture proposed by \cite{azulay2022multigrid}. Furthermore, we introduce a data-efficient light-weight retraining process to maintain the network performance throughout the inversion process and reduce the overall computational complexity and solve-time of the inversion process.

\section{Preliminaries and Background}

\subsection{The Full Waveform Inversion problem}
Full waveform inversion (FWI) is a challenging inverse problem in which we estimate
the wave propagation velocity of a given medium using observed wavefield data on the
boundaries. We consider the frequency domain version of FWI. 

\paragraph{The forward problem: the Helmholtz equation.} To model the waveforms, we consider the discrete acoustic Helmholtz equation as a forward problem, assuming a constant density media
\begin{equation}\label{eq:helmholtz}
   -\Delta u(x,\omega)-\omega^2m(x)(1-\gamma i)u(x,\omega)=g(x,\omega), \quad  x\in \Omega.
\end{equation}
The unknown $u(x,\omega)$ represents the pressure field in the frequency domain, $\Delta$ is the
Laplacian operator, $m(x)$ is the heterogeneous squared slowness of the medium which is the physical parameter we wish to estimate in this work. That is $m(x) = 1/c(x)^2$ where $c(x)$ is the acoustic wave velocity, which varies with position, and $\omega=2\pi f$ denotes angular frequency, a
scalar measure of rotation rate ($f$ is the frequency, measured in Hertz in our context). The source term is denoted by $g$, and is assumed to be a discretization of a delta function, $\delta(\hat{x}-x)$ located at $x$. The parameter $\gamma$ indicates the fraction of attenuation (or damping) in the medium, and can possibly be considered as varying attenuation, i.e., $\gamma(x)$. $i =\sqrt{-1}$ is the imaginary unit. Free space or absorbing (non-reflecting) boundary conditions are applied at the boundaries $\Gamma=\partial\Omega$.

%\tred{You need to move from a continuous description to a discrete notation ($u$ vs $\mathbf{u}$), and say that you have a linear system after discretization. Also you need to finalize this such that the solution is $\bfu(\bfm,\omega,x)$.}

Equation \eqref{eq:helmholtz} is given in continuous space, and to solve it we discretize it on a grid. We
use the finite difference discretization, obtained by a standard central difference scheme  which leads to a system of linear equations
\begin{equation}\label{eq:linear_helmholtz}
    \mathcal{H}(\bfm,\omega)\bfu(\bfx,\omega) = \bfg(\bfx,\omega).
\end{equation}
Here, boldface letters, e.g., $\bfu$, denote the functions sampled on the grid. $\mathcal{H}$ is the linear operator defined by the finite difference stencil of the Helmholtz equation. When discretizing the Helmholtz equation we must use a small enough grid size such a wavelength is accomodated on several grid nodes, depending on the discretization order (typically, 10 grid nodes per wavelength are used). This typically requires a fine mesh for high wavenumbers, making the grid and system $\mathcal{H}\bfu = \bfg$ above huge, involving possibly hundreds of millions of unknowns.

\paragraph{The inverse problem.} In FWI, sources are located at many locations on the top part of the domain, and the waveform that is generated by each source is recorded at the locations where receivers are placed. We denote a discrete solution of \eqref{eq:linear_helmholtz} as $\bfu(\bfm,\omega,x_s)$ for $\bfg$ that is a discretization of a source at location $x_s$. In our formulation, each observed data sample corresponding to a source $x_{s}$ and frequency $\omega_{j}$, is assumed to be given by
\begin{equation}\label{eq:obs_sampling}
\bfd^{\text{obs}}(\bfm_{\text{true}},x_{s}, \omega_{j}) = \bfP_{s}^{\top}\bfu(\bfm_{\text{true}},\omega_{j},x_{s}) + \epsilon_{sj}
\end{equation}
where $\bfP_{s}^{\top}$ is a sampling matrix that measures the wave field $\bfu$ that is generated by the source at $x_{s}$ at the locations of the receivers, and $\epsilon_{sj}$ is noise, which we assume to be i.i.d sampled from a Gaussian distribution with zero mean and variance $\Sigma_{sj}$. Given data for many sources and multiple frequencies, we aim to estimate the true underlying velocity model, $\textbf{m}_{\text{true}}$, by minimizing the difference between the measured and simulated data.

This may be done by solving the following PDE-constrained optimization problem
\begin{equation}\label{eq:fwi_opt}
\begin{aligned}
    & \underset{\bfm, \{\bfu_{sj}\}_{s=1,j=1}^{n_{s},n_{f}}}{\arg\min} \Phi(\textbf{m},\{\textbf{u}_{sj}\})=\sum_{j=1}^{n_{f}}\sum_{s=1}^{n_{s}} \Vert
    \bfP_{s}^{\top}\bfu_{sj} - \bfd_{sj}^{\text{obs}} \Vert_{\Sigma_{sj}^{-1}}^{2} + \alpha \mathcal{R}(\bfm) \\
    & \text{s.t. } \mathcal{H}(\bfm,\omega)\bfu_{sj} = \bfg_{s}
\end{aligned}
\end{equation}
where $\bfu_{sj}$ is the simulated waveform for source $x_{s}$ and frequency $\omega_{j}$, which is predicted for a given velocity model $\bfm$, according to the forward problem defined in \eqref{eq:helmholtz}, the data terms $\textbf{d}_{sj}^{\text{obs}}$ are the corresponding observed data as in \eqref{eq:obs_sampling}. Due to the fact that we assume $\bfm$ to be layered model we include a regularization term, $\mathcal{R}(\bfm)$, which promotes piece-wise smooth functions like total variation regularization \cite{rudin1992nonlinear}. Later, we detail the regularization terms we use in our experiments).

There are several approaches for solving constrained optimization problem such as \eqref{eq:fwi_opt}. One of which is the Lagrange multipliers approach, where a Lagrange variable will be introduced to each pair $s,j$ \cite{haber2001preconditioned,metivier2017full}. Another common approach, which we follow in this paper, is to eliminate the PDE constraint and include it in the optimization formulation itself. Resulting in the unconstrained formulation
\begin{equation}\label{eq:fwi_unconstrained}
    \underset{\textbf{m}}{\arg \min} \,\,\Phi(\bfm)=\sum_{j=1}^{n_{f}}\sum_{s=1}^{n_{s}} \Vert
    \bfP_{s}^{\top}\mathcal{H}(\bfm,\omega_{j})^{-1}\bfg_{s} - \bfd_{sj}^{\text{obs}} \Vert_{\Sigma_{sj}^{-1}}^{2} + \alpha \mathcal{R}(\bfm).
\end{equation}

%However, in this formulation the optimization is forced to fulfill the PDE constraints at all time, thus resulting in highly non-linear process. \cite{van2015penalty} suggested to relax these constraints by allowing the terms $\bfu_{sj}$ to not fully satisfy the constraints and add a penalty term on the constraints error 

%\tred{ERAN: I am not sure we really need the equation below}.
%
%\begin{equation}\label{eq:fwi_unconstrained_penalty}
%\begin{split}
% \underset{\textbf{m}, \{u_{sj}\}_{s=1,j=1}^{n_{s},n_{f}}}{\arg \min} \Phi(\textbf{m},\{u_{sj}\}) &=\sum_{j=1}^{n_{f}}\sum_{s=1}^{n_{s}} \Vert
 %   \textbf{P}_{s}^{\top}\textbf{u}_{sj} - \textbf{d}_{sj}^{\text{obs}} \Vert_{\Sigma_{sj}^{-1}}^{2} + \\
  %  & \beta\sum_{j=1}^{n_{f}}\sum_{s=1}^{n_{s}}\Vert \mathcal{H}(\kappa_{\textbf{m}},\omega_{j})\textbf{u}_{sj} - g_{s} \Vert_{2}^{2}+\alpha R(\textbf{m})
%\end{split}
%\end{equation}

%These approaches, the Lagrange multipliers and the penalized relaxed unconstrained formulation include iterative updates of $\{\bfu_{sj}\}$, which require a lot of memory and are almost not feasible when dealing with large-scale data or in 3D settings. \tred{ERAN: this is not our battle here - it was Sagi's... we need to be less offensive.}

%\input{pages/paper_FWI_background}
%% NEED TO think about shifted laplacian at the end for the paper

\subsection{The Simultaneous Sources Approach} \label{sec:simsrc} A computationally efficient approach for solving \eqref{eq:fwi_unconstrained} utilizes the concept of \textit{simultaneous sources}, which is obtained by trace estimation \cite{haber2012effective}. That is, given a matrix $\bfA\in \mathbb{R}^{m \times n}$ we can approximate its Forbenius norm by \cite{avron2011randomized}
\begin{equation}\label{eq:trace_estimation_def}
   \Vert \bfA \Vert_{F}^{2} = trace(\bfA^{T}\bfA) = \mathbb{E}_{x}\Vert \bfA\bfx \Vert_{2}^{2} \approx \frac{1}{p}\sum_{i=1}^{p}\Vert \bfA\bfx_{i}\Vert_{2}^{2} = \frac{1}{p}\Vert \bfA\bfX \Vert_{F}^{2}
\end{equation}
where each $\bfx_{i}\in \mathbb{R}^{n}$ is chosen from Rademacher distribution and $\bfX$ is $n \times p$ matrix whose $i$-th column is $\bfx_{i}$. This approximation allows us to approximate the norm of a large $m\times n$ matrix using an $m\times p$ matrix, which can significantly reduce computational cost when $p\ll n$.

Utilizing \eqref{eq:trace_estimation_def} for approximating the data term in \eqref{eq:fwi_unconstrained} is only possible when $\textbf{P}_{s}^{\top}$ and $\Sigma_{sj}$ are not source dependent, i.e., when all the receivers record the waveform from all the sources. To mitigate this issue, \cite{liu2018simultaneous} introduces a new set of variables $\hat{\textbf{d}}_{sj}$ that are defined over all the locations of the receivers for all the sources, and an additional penalty term is added to ensure that $\hat{\textbf{d}}_{sj}$ is similar to the original $\textbf{d}_{sj}^{\text{obs}}$ at the locations of the receivers that actually record the waveform for each source $s$. Thus, \eqref{eq:fwi_unconstrained} can be formulated as
\begin{equation}\label{eq:fwi_unconstrained_dhat}
        \underset{\bfm, \{\hat{\bfd}_{sj}\}_{s=1,j=1}^{n_{s},n_{f}}}{\arg \min} \Phi(\bfm)=\sum_{j=1}^{n_{f}}\sum_{s=1}^{n_{s}} \Vert
    \bfP^{\top}\mathcal{H}(\bfm,\omega_{j})^{-1}\bfg_{s} - \hat{\bfd}_{sj} \Vert_{\hat\Sigma_{j}^{-1}}^{2} + 
     \eta\Vert \hat{\bfP}_{s}^{\top}\hat{\bfd}_{sj} - \bfd_{sj}^{\text{obs}}\Vert_{\Sigma_{sj}^{-1}}^{2}
     + \alpha \mathcal{R}(\bfm)
\end{equation}
where $\bfP$ is an operator that projects a vector onto the union of the receivers’ locations, and $\hat{\bfP}_{s}$ chooses the subset of receivers for source $s$ out of that union, i.e., $\bfP_{s}^{\top} = \hat{\bfP}_{s}\bfP^{\top}$. To apply the simultaneous sources technique to \eqref{eq:fwi_unconstrained_dhat} we define a $n_{s}\times p$ matrix $\bfX$ from Radermacher distribution, and by \eqref{eq:trace_estimation_def} we obtain
\begin{equation}\label{eq:fwi_data_term_approximation}
\begin{aligned}
    \sum_{s=1}^{n_{s}} \Vert
    \bfP^{\top}\mathcal{H}(\bfm,\omega_{j})^{-1}\bfg_{s} - \hat{\bfd}_{sj} \Vert_{\Sigma_{sj}^{-1}}^{2} &= \Vert
    \bfP^{\top}\mathcal{H}(\bfm,\omega_{j})^{-1}\bfG - \hat{\bfD}_{j} \Vert_{\Sigma_{j}^{-1}}^{2}\\
    &\approx \frac{1}{p}\Vert
    \bfP^{\top}\mathcal{H}(\bfm,\omega_{j})^{-1}\bfG\bfX - \hat{\bfD}_{j}\bfX \Vert_{\Sigma_{j}^{-1}}^{2}
\end{aligned}
\end{equation}
This effectively reduces the number of sources to $p$ in every iteration.

\subsection{Frequency continuation and Gauss-Newton}
Since FWI is ill-posed and highly non-convex, local minima can pose major obstacles when solving FWI by optimization. That is, converging to a local minimum may lead to an non-plausible estimated slowness model. Therefore, avoiding local minima can be done effectively by using the method of frequency continuation (FC) \cite{pratt1998gauss,pratt1999seismic}. As shown in algorithm \ref{alg:frequency_continuation}, the FC algorithm iterates over ranges of frequencies, called cycles, each defined by an initial and final frequency.
In each such cycle, we solve the optimization problem for the given frequencies in a sliding window manner using the model estimated by the previous cycle as an initial guess for the current cycle. Solving the optimization problem with respect to $\mathbf{m}$ is done iteratively using Gauss-Newton (GN) \cite{pratt1998gauss}. Specifically, at each iteration $t$ of GN we obtain the linear approximation
\begin{equation}\label{eq:GN_linear_approximation}
    \bfu_{sj}(\bfm^{(t)}+\delta\bfm) = \bfu_{sj}(\bfm^{(t)}) + \bfJ_{sj}(\bfm^{(t)})\delta\bfm
\end{equation}
where $\bfu_{sj}(\bfm^{(t)}) = \mathcal{H}(\bfm^{(t)},\omega_{j})^{-1}(\bfg_{s})$ and the Jacobian matrix $\bfJ_{sj}(\bfm^{(t)})$ is defined as
\begin{equation}\label{eq:GN_J}
    \bfJ_{sj}(\bfm^{(t)}) = \nabla_{\bfm}\bfu_{sj}=-\omega_{j}^{2}\,\mathcal{H}(\bfm,\omega_{j})^{-1}\text{diag}(\mathcal{H}(\bfm,\omega_{j})^{-1}\bfg_{s})
\end{equation}
\begin{algorithm}
 \caption{Frequency Continuation}
 \textbf{Input:}\\
 $\omega_{1}< \ldots <\omega_{n_{f}}$: frequencies\\
 $ws$: window size\\
  $\mathcal{C}=[(i_{start}^{(1)},i_{end}^{(1)}),\dots, (i_{start}^{(k)},i_{end}^{(k)})]$: set of cycles\\
  \textbf{Algorithm:}\\
  \For{$(i_{start},i_{end})\in\mathcal{C}$}
  {
  Initialize $\textbf{m}$ (or use the model from previous cycle)\\
  \For{$i=i_{start}:i_{end}$}{
  $\textbf{m}\leftarrow$ Approximately solve $\Phi$ using data for the frequencies.\\ 
  $\omega_{\max\{i-ws,1\}}\ldots \omega_{i}$, starting from the previous model $\textbf{m}$.
 }
 }
 \label{alg:frequency_continuation}
\end{algorithm}

At each iteration we solve the forward problem \eqref{eq:helmholtz} for each pair of frequency and source, whether encoded through the simultaneous sources method or not. At each GN iteration one has to solve the linear system $\bfH\delta\bfm = -\nabla_{\bfm}\Phi(\bfm^{(t)})$, where $\bfH$ is the Gauss-Newton Hessian and is defined by
\begin{equation}\label{eq:GN_Hessian}
    \bfH = \sum_{s,j}\bfJ_{s,j}(\bfm^{(t)})^{\top}\bfP\bfP^{\top}\bfJ_{s,j}(\bfm^{(t)}) + \alpha\nabla^{2}\mathcal{R}(\bfm^{(t)}). 
\end{equation}
The system here is solved by the Conjugate Gradient algorithms, involving many applications of the Hessian to vectors, which requires solving the forward and the adjoint systems involving $\mathcal{H}$ and $\mathcal{H}^{*}$, respectively.  We note that the right-hand-side vectors for the Jacobians are arbitrary (that is, not the source functions) as they come from the optimization process.

\subsection{Shifted Laplace multigrid-augmented CNN preconditioner}
\begin{figure}
\centering
\includegraphics[width=0.8\textwidth]{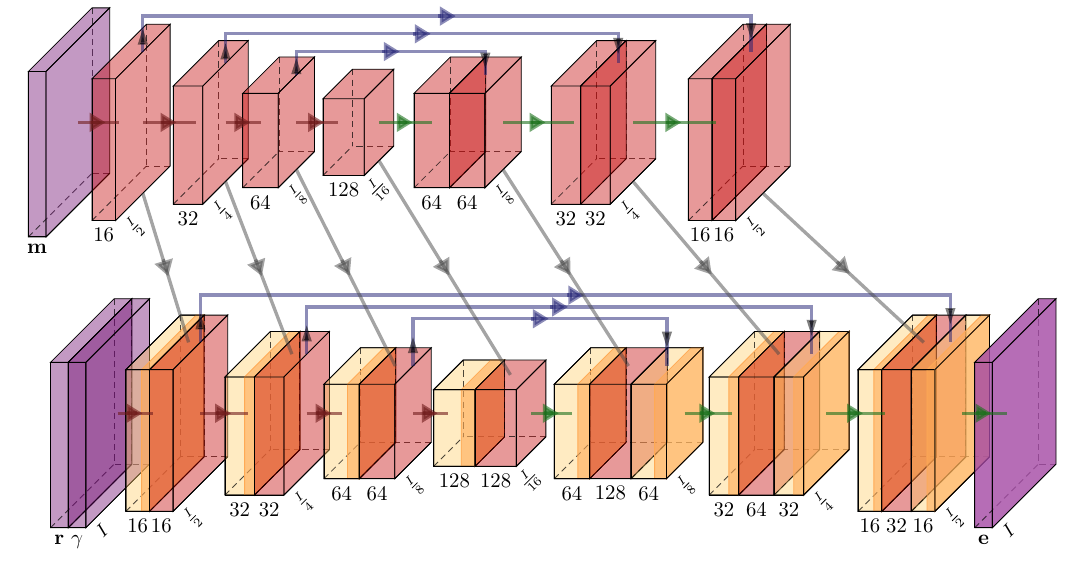}
\caption{\footnotesize\textit{The Encoder-Solver architecture. The Encoder receives the slowness (squared) and progressively coarsens the feature map using a convolutional layer followed by 2 consecutive ResNet layers. On the way up we up-sample the coarse vector using a convolutional layer, concatenate the feature vector of the corresponding size, and apply a ResNet layer. The Solver network maps a residual vector $\bfr$ to an error $\bfe$. The Solver integrates the learned feature vector of the Encoder at each level.}}
\label{fig:encoder_solver_architecture}
\end{figure}

The Shifted Laplace operator in \cite{erlangga2006novel} was proposed to allow the solution of the Helmholtz equation using multigrid, in a Krylov subspace method (e.g. GMRES). The preconditioner is given by:
\begin{equation}\label{eq:SL_operator}
 P \bfu = -\Delta_h \bfu-\omega^2\bfm(\bfx)(\alpha-\beta i)\bfu, \quad \alpha,\beta \in \mathbb{R}
\end{equation}
where typically, $\alpha=1$ to maintain similarity to $\mathcal{H}$, and $\beta=0.5$ which was shown in \cite{erlangga2006novel} to lead to an efficient and robust preconditioning operator. However, the SL method (similar to relaxation methods) efficiently handles error modes with rather large absolute eigenvalues, as the shift does not perturb those eigenvalues by a significant factor. Therefore, \cite{azulay2022multigrid} introduced a CNN-based augmentation to the SL preconditioner, such that the augmentation focuses on "algebraically smooth" errors. The augmentation is obtained by an Encoder-Solver model, depicted in Figure \ref{fig:encoder_solver_architecture}, responsible for mapping residual vectors generated by a Krylov method to the corresponding error vectors, which is then followed by a multigrid cycle \cite{calandra2013improved}. 

The Encoder-Solver model consists of two separate U-Net models. The Encoder U-Net receives only the squared slowness model $\bfm$ as input and generates context vectors
\begin{equation}\label{eq:encoder}
\{\bfz_i\} = \text{EncoderNet}(\bfm;\boldsymbol{\theta}_{e}), 
\end{equation}
aiming to generalize solely over the slowness model. The Solver U-Net integrates the learned feature maps $\{\bfz_i\}$ from the encoder when generalizing over the residual vectors. This way the solver network has more information on the slowness model, and it allows us to compute the feature maps $\{\bfz_i\}$ of a given slowness model only once, and solve for any right-hand-side residual vector. This property is highly desirable in the case of FWI where the Helmholtz equation is solved for numerous right-hand-side vectors for a given slowness model. Overall, the preconditioning process, denoted as $M_{VU}$,  is given by
\begin{equation}\label{eq:vu_preconditioner}
 M_{VU}(\textbf{r}) = \text{V-cycle}(\textbf{e}_{0}=\text{SolverNet}(\textbf{r},\gamma;\{\bfz_i\},\boldsymbol{\theta}_s), \textbf{r}, \text{levels}=3)
\end{equation}
acting on a residual vector $\bfr$ generated by a Krylov method.

\section{Method}
Our method incorporates the U-Net Encoder-Solver preconditioner of the Helmholtz equation in the inversion process, both in the data simulations and in the sensitivity (Jacobian) computations in equation \eqref{eq:GN_J}. The CNN network in \cite{azulay2022multigrid} was proposed in order to solve general Helmholtz problems, and hence, was trained over a general data set of slowness models for a particular setting of the problem (e.g., grid-size, domain, and proportion between the horizontal and vertical grid sizes $h_x$ and $h_y$). Any deviation from these settings can hinder the effectiveness of the approach. Furthermore, it is shown that the performance of the model deteriorates when solving an equation corresponding to an out-of-distribution slowness model. Namely, the
generalization over the slowness model hurts the performance. It is also shown that briefly re-training the network on the slowness model of interest reconstitutes accuracy for this model. 

%Furthermore, the solver is trained over a data set of slowness models. In this context, it is shown that if the slowness model is not part of the distribution of the data set, the performance deteriorates. It is also shown that the solver is much better at solving an equation when trained on the same slowness model that corresponds to the equation. In other words, the generalization over the slowness model hurts the performance. 

Here, we wish to exploit the fact that we solve an inverse problem to estimate a single (yet, unknown) slowness model $\bfm(\bfx)$ for a given problem setting (e.g., domain, grid-size), and that we know the initial guess of the inversion process. We build on the fact that the model is built gradually and does not change dramatically between the frequency continuation iterations. Hence, to cope with the gradual change in the underlying slowness model we introduce and integrate a light-weight retraining procedure of the Encoder-Solver model in between the inversion iterations so that its weights are updated together with the inversion to solve the forward problems well.

In summary, we employ the Encoder-Solver model as a preconditioner to the Helmholtz equation, where the model has been trained to consume general right-hand-sides (residuals) generated by a Krylov method until convergence. 
The advantages of this approach in our context  are as follows:
\begin{enumerate}
\item Since there are many sources in typical FWI scenarios, we wish to train one CNN model for all sources. Also, the CNN model should allow us to solve the discrete Helmholtz equation to any desired accuracy as the estimation of deep reflectors requires quite accurate simulations. Accurate solutions are also required for a consistent gradient computation in \eqref{eq:GN_J}. Applying the CNN as a preconditioner allows us to reach any desired accuracy. 
\item It is highly preferable that the model can solve problems with general right-hand sides. This is required for the multiplication of the Jacobian in \eqref{eq:GN_J} and its transpose with a vector when computing the gradient or when solving the inner Newton problem with CG. Furthermore, the forward solutions for random sources are also needed for the data simulation in the simultaneous sources method detailed in section \ref{sec:simsrc}, where in each iteration we encode the sources using a random matrix. 
\end{enumerate}

%\tred{The following paragraph can move to 4.1 or even deleted.} The U-Net architecture is a convolutional neural network that is able to capture and localize significant contexts and features using two rather symmetric parts - contraction and expansion \cite{ronneberger2015u}. Moreover, the U-Net architecture is similar in spirit to the Multigrid V-cycle method and shares common properties. Thus, the Encoder-Solver can be efficiently used as a preconditioner component in an iterative solver of the descretized Helmholtz equation \cite{azulay2022multigrid}. The given slowness model is encoded using the Encoder U-Net prior to the solution process. Then, the Solver network is applied at each iteration, with the readily available feature vectors inside it. 

We further elaborate on the training process of the Encoder-Solver model and the data generation scheme in Section \ref{subsecction:data_generation_training}. Next, in Section \ref{subsection:retraining} we present our retraining integrated frequency continuation procedure. Lastly, in Section \ref{subsection:adjoint_system} we show how the Encoder-Solver model is able to solve the adjoint system efficiently.

\subsection{Training and data generation}\label{subsecction:data_generation_training}
\paragraph{Training.} The training of the Encoder-Solver model is performed in a supervised manner. Specifically, let
\begin{equation}
    \bfe^{\text{net}}(\bfr,\bfm,\gamma;\boldsymbol{\theta}) = \text{SolverNet}(\bfr,\gamma; \text{EncoderNet}(\bfm;\boldsymbol{\theta}_e),\boldsymbol{\theta}_s)
\end{equation}
be a forward application of the network for a given residual vector $\bfr$, slowness (squared) model  $\bfm(\bfx)$ and attenuation model $\gamma(\bfx)$ (which is the same for every sample in the dataset). $\boldsymbol{\theta}$ denotes the set of trainable weights of both networks $\{\boldsymbol{\theta}_e,\boldsymbol{\theta}_s\}$. We seek to minimize the mean squared error (MSE)
\begin{equation}\label{eq:mse}
    \underset{\boldsymbol{\boldsymbol{\theta}}}{\min}\,\frac{1}{m}\sum_{i=1}^{m}\Vert \bfe^{\text{net}}(\bfr_{i},\bfm_{i},\gamma;\boldsymbol{\theta}) - \bfe_{i}^{\text{true}}\Vert_{2}^{2}
\end{equation}
for each batch of data samples $\{(\bfe_{i}^{\text{true}}, \bfr_{i},\bfm_{i},\gamma)\}_{i=1}^{m}$, where $\mathcal{H}_{i}^{h}\bfe_{i}^{\text{true}}=\bfr_{i}$. $\mathcal{H}_{i}^{h}$ is the Helmholtz equation descretized operator as defined in \ref{eq:linear_helmholtz} for the given slowness model $\bfm_{i}(\textbf{x})$. The minimization is performed by a stochastic gradient descent optimizer, wherein we sweep through all the training examples in batches each epoch. We use the ADAM optimizer \cite{kingma2014adam} with learning rate decay.

\paragraph{Data Generation.} Creating the dataset for training our network requires creating data samples of the form $(\bfe_{i}^{\text{true}}, \bfr_{i},\bfm_{i},\gamma)$. Where $( \bfr_{i},\bfm_{i},\gamma)$ is the input to the network and $\bfe_{i}^{\text{true}}$
 is the corresponding error vector used to compute the misfit term in \eqref{eq:mse}. $\gamma(\bfx)$ is an absorbing layer (ABL) \cite{engquist1977absorbing,engquist1979radiation,erlangga2006novel} that we use to model open domains in \eqref{eq:helmholtz}. Specifically, in our case, $\gamma(\bfx)$ is a two dimensional function that goes from $0$ to $1$ towards the boundaries, except for the top boundary, which indicates the sea layer, which we allow reflections from. $\{\bfm_{i}(\bfx)\}$ are spatially dependent random linear slowness (squared) models (e.g., see Fig. \ref{fig:seg_initial_model}) and are generated with correspondence to the initial guess used for the FWI optimization process. That is, we randomly select physically applicable values for the top and bottom layers, and the slowness model is obtained by uniform intervals between these two values. Lastly, since the relation between the error vector and the residual vector is described as a linear system, $\mathcal{H}_{i}^{h}\bfe_{i}^{\text{true}}=\bfr_{i}$, the creation of $\bfr_{i}$ and $\bfe_{i}^{\text{true}}$ pairs is straightforward, and does not involve solving the PDE directly to obtain ground-truth solution. However, because we wish to focus on algebraically smooth errors (errors with low residuals), we adapt our data by applying a random number of GMRES iterations with a V-cycle preconditioner. That is, we first generate a random vector $\bfx_{i}$ from a complex normal distribution, which is in correspondence to the random right-hand-sides vectors we solve for throughout the optimization process. Then we compute the right-hand-side vector $\bfb_{i} = \mathcal{H}_{i}^{h}\bfx_{i}$ and apply
\begin{equation}\label{eq:x_tilde_data_generation}
    \Tilde{\bfx}_{i} = \text{GMRES}(\mathcal{H}_{i}^{h},M=\text{V-cycle},\bfb_{i},\bfx^{(0)}=\textbf{0},iter\in\{1,\dots,10\}).
\end{equation}
Following that, we compute $\bfr_{i} = \bfb_{i} - \mathcal{H}_{i}^{h}\Tilde{\bfx}_{i}$ and $\bfe_{i}^{\text{true}} = \bfx_{i} - \Tilde{\bfx}_{i}$.

\subsection{Retraining integrated frequency continuation }\label{subsection:retraining}
Here we describe the key idea for utilizing the Encoder-Solver described above for solving FWI most effectively. The problem with the CNN-based solver above is that it cannot be highly effective on all slowness models, but given one slowness model and proper training, it performs well. We wish to exploit this property through mini-retraining sessions of the solver against the gradually evolving inversion iterates as detailed below. 

In our method, we solve FWI using the frequency continuation method in Algorithm \ref{alg:frequency_continuation}. As the algorithm progresses the slowness model is gradually changed to fit the observed data. Thus, the slowness model gradually deviates from the initial linear model distribution that our solver was trained on. As a result, the Encoder-Solver model is required to generalize over unseen slowness models, which leads to poor performance. However, when a light-weight retraining procedure is applied, the performance of the model is kept and the iteration count needed for the solution of the Helmholtz equation remains low. To this end, we incorporate a retraining procedure when a new slowness model is obtained, i.e., after each inner iteration within the frequency continuation algorithm. 

At this point, clearly, there is a trade-off between the cost of the retraining procedure and the improvement of our solver network. The more resources we invest in the training procedure the more our solver is able to adapt to the new slowness model. However, our main goal is to control the computational resources used throughout the inversion process, which is one of the main obstacles of FWI. In addition to the cost of the training itself a major part of the computational cost is due to the process of data creation, which is described in Subsection \ref{subsecction:data_generation_training}. Therefore, our proposed retraining procedure is data efficient and computationally light, and is described in Algorithm \ref{alg:retraining}. 

\begin{algorithm}
\caption{Light-weight Retraining Procedure}
\LinesNumbered
\textbf{Input:}\\
\textit{epochs}: total number of epochs.\\ 
\textit{data\_epochs}: number of initial epochs that also generate data.\\
\textit{initial\_size}: the size of the initial data set.\\
\textbf{Algorithm:}\\
Create an initial dataset $\mathcal{D}$ corresponding to $\bfm$ of size \textit{initial\_size}\\
 \For{$i=1:epochs$}{
  Train model on $\mathcal{D}$\\
  \If{$i\leq data\_epochs$}
        {    \ForEach{$(\bfr,\bfe,\gamma,\bfm)\in\mathcal{D}$}
        {
          $\bfe_{\text{new}}\leftarrow \text{Encoder-Solver}(\bfr,\bfm,\gamma)$\\
          %$\bfr_{\text{new}}\leftarrow \mathcal{H}^{h}\bfe_{m}$
          $\Tilde{\bfe}\leftarrow \text{GMRES}(\mathcal{H}^{h}, M=\text{V-cycle},\bfr,\bfx^{(0)}=\bfe_{\text{new}}, iter=1)$\\
          $\Tilde{\bfr}\leftarrow \mathcal{H}^{h}\Tilde{\bfe}$\\
          Append $(\Tilde{\bfr},\Tilde{\bfe},\gamma,\bfm)$ to $\mathcal{D}$
        }
    }
 }
 \label{alg:retraining}
\end{algorithm}

To start the algorithm we create a small dataset, and the rest of the data samples are gradually created throughout the retraining process in a manner that is compatible with our solver inaccuracies. That is, for several initial epochs we re-use the network outputs obtained in the training as new initial guesses in the process of creating a new $(\bfr,\bfe)$ corrected pair. The process resembles ideas presented in \cite{fung2022jfb}, where the network is used to generate more data to obtain a fixed-point process of implicit (or, recurrent) neural networks. Specifically, in our method we initially create dataset of size $initial\_size = 128$ and gradually create the rest of the data samples for the first $data\_epochs=3$ epochs, resulting in a final dataset of $1024$ samples. Then, we continue to train our model for the rest of the epochs until reaching our epoch limit of 30. Note that in the data creation epochs we obtain a new error vector using GMRES, but with one iteration only, which help reduce the computational cost and still improve our Encoder-Solver preconditioner performance. Moreover, the final dataset size (1024) is considered small compared to the initial dataset size that was used to train the network, keeping the retraining process at each iteration computationally light-weight.

\subsection{The adjoint system}\label{subsection:adjoint_system}
Throughout the inversion process, one also has to solve the adjoint system involving $\mathcal{H}^{\ast}$, as shown in \eqref{eq:GN_Hessian} and \eqref{eq:GN_J}. However, the Encoder-Solver networks are trained on data created by the system obtained by $\mathcal{H}$, and we do not wish to train two sets of networks to support both the standard and adjoint Helmholtz systems. Thus, in order to adapt our solver for the adjoint system we alter the right-hand-side vector of the system rather than training the Encoder-Solver on more data or have two sets of networks. That is, suppose we have the adjoint system
\begin{equation}\label{eq:adjoint_system}
    \mathcal{H}^{\ast}\bfx=\bfb
\end{equation}
we can re-write it with the real and imaginary terms, resulting in the equation
\begin{equation}\label{eq:adjoint_system_parts}
    \left(\text{re}(\mathcal{H}^{\top}) - \text{im}(\mathcal{H}^{\top})i \right)(\text{re}(\bfx)+\text{im}(\bfx)i) = \text{re}(\bfb)+\text{im}(\bfb)i
\end{equation}
where $\text{re}(\cdot), \text{im}(\cdot)$ represent the real and imaginary parts, respectively, and $i$ is the imaginary unit. The Helmholtz operator matrix $\mathcal{H}$ has symmetric real and imaginary parts, i.e., $\text{re}(\mathcal{H}^{\top})=\text{re}(\mathcal{H})$ and $\text{im}(\mathcal{H}^{\top})=\text{im}(\mathcal{H})$. Thus, equation \eqref{eq:adjoint_system_parts} translates to
\begin{equation}\label{eq:helmholtz_system_parts}
    \left(\text{re}(\mathcal{H}) - \text{im}(\mathcal{H})i \right)(\text{re}(\bfx)+\text{im}(\bfx)i) = \text{re}(\bfb)+\text{im}(\bfb)i
\end{equation}
and we obtain a system of equations, for the real and imaginary part, separately
\begin{equation}
    \begin{aligned}
        &(1)\quad \text{re}(\mathcal{H})\text{re}(\bfx)+\text{im}(\mathcal{H})\text{im}(\bfx) = \text{re}(\bfb)\\
        &(2)\quad\text{re}(\mathcal{H})\text{im}(\bfx)-\text{im}(\mathcal{H})\text{re}(\bfx) = \text{im}(\bfb).
    \end{aligned}
\end{equation}
Now, performing $(1)-i(2)$ yields the equation
\begin{equation}\label{eq:Helmholtz_after_alter}
    \mathcal{H}\bfx^{\ast} = \bfb^{\ast}.
\end{equation}
which includes the standard $\mathcal{H}$ with conjugate solution and right-hand-side. Therefore, the result for the original adjoint system presented in \eqref{eq:adjoint_system} is the conjugate of the vector obtained by solving \eqref{eq:Helmholtz_after_alter}, i.e., $(\bfx^{\ast})^{\ast}=\bfx$, as required.
Since our Encoder-Solver model was trained on random right-hand-side vectors obtained by the operator $\mathcal{H}$, equation \eqref{eq:Helmholtz_after_alter} is compatible with our solver. 
\section{Numerical Results}
In this section, we report our results and demonstrate
the efficiency of the proposed technique, focusing mainly on the improved performance of our forward solver network throughout the inversion process. In our experiments, we perform FWI with simultaneous sources only, and we do not augment the inversions with other modalities or techniques known in the literature. In principle, such techniques can be used in addition to our approach in more complicated real-life experiments. We conduct our experiments on the 2D SEG/EAGE salt model \cite{aminzadeh19973} and on the Marmousi model \cite{brougois1990marmousi}. We compare the iteration count and solve time for solving the FWI problem with and without our proposed retraining procedure. All of our experiments and time measurements were conducted using NVIDIA RTX-3090 GPU operated by the Julia Flux framework.

\textbf{Initial training details:} We train our network on a dataset consisting of $20,000$ samples, and a validation set of $1,000$ samples, using grid size of $608\times 304$. We train for $120$ epochs, the batch size is set to $16$ and the initial learning rate is set to $10^{-4}$.

\textbf{Retraining details:} We perform the retraining procedure after each window in the frequency continuation algorithm \ref{alg:frequency_continuation} (after each inner iteration). Therefore, we train the Encoder-Solver model using data samples with grid size that corresponds to the highest frequency of the window. Namely, grid size that obtains 10 grid points per wavelength. This further reduce the computational cost of the retraining procedure while assuring the accuracy of the Encoder-Solver model.

\subsection{Regularization terms used for the reconstruction}
The objective functions in the FWI formulations in equations \eqref{eq:fwi_opt}, \eqref{eq:fwi_unconstrained}, and \eqref{eq:fwi_unconstrained_dhat} contain a regularization term $\mathcal{R}(\bfm)$. Based on the work \cite{treister2017full} we apply two regularization
functions in our experiments. The first of which is a high order regularization called spline smoothing, given by
\begin{equation}\label{eq:high_order_regularization}
    \mathcal{R}_{high}(\bfm) = \Vert \Delta_{h}(\bfm - \bfm_{ref}) \Vert_{2}^{2}
\end{equation}
The goal of this regularization is to create a smooth model from high-frequency data.
The reference model $\bfm_{ref}$ is set to be the initial guess for the inversion, and while
using this regularization, we keep $\bfm_{ref}$ fixed. We use \eqref{eq:high_order_regularization} to obtain a good smooth
model, so that the rest of the process will result in a plausible reconstruction and avoid local minima. The second regularization function is a standard diffusion regularization
\begin{equation}\label{eq:diff_regularization}
    \mathcal{R}_{diff}(\bfm) = \Vert \nabla_{h}(\bfm - \bfm_{ref}) \Vert_{2}^{2}
\end{equation}
where $\nabla_{h}$ represents a discrete gradient on a nodal grid. Using \eqref{eq:diff_regularization} results in a rather sharp reconstruction and is suitable after a smooth model is constructed.

\subsection{SEG/EAGE salt model setup and results} 
\begin{figure}[H]
    \centering
    \begin{subfigure}{.48\textwidth}
        \includegraphics[width=\textwidth]{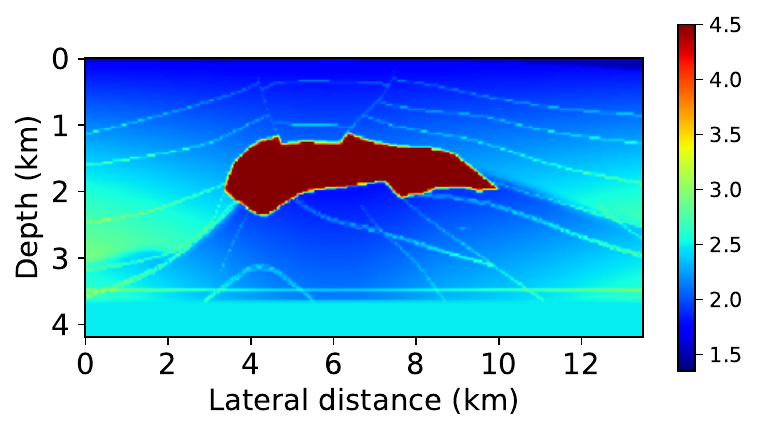}
		\caption{True velocity model}
        \label{fig:seg_true_model}
	\end{subfigure}
    \hspace*{\fill}
    \begin{subfigure}{.48\textwidth}
        \includegraphics[width=\textwidth]{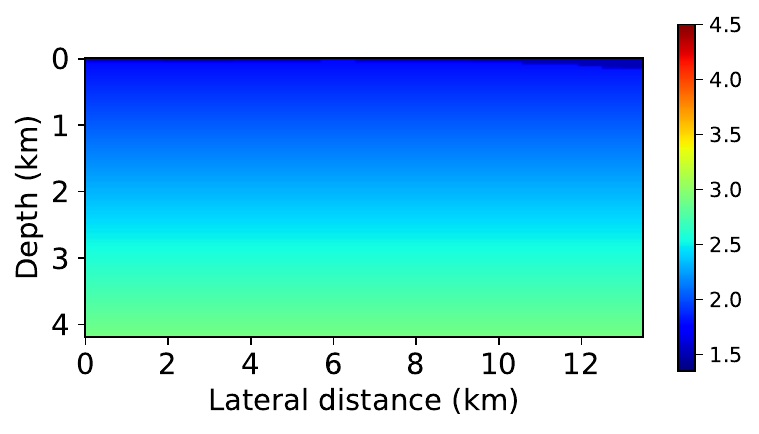}
		\caption{Reference initial model}
        \label{fig:seg_initial_model}
	\end{subfigure}      
\caption{\footnotesize\textit{The 2D SEG/EAGE salt velocity model and initial guess}}
\end{figure}

Our first set of experiments is conducted using
the SEG/EAGE model presented in Fig. \ref{fig:seg_true_model}. The model is described
by a $608 \times 304$ grid, representing an area of size $13.5km \times 4.2km$. The data
is generated by first solving \eqref{eq:helmholtz} for frequencies $\omega_{j} = 2\pi f_{j}$, where
\begin{equation}  
\{f_{j}\} = \{\underbrace{2.6, 2.9, 3.2, 3.5}_{low}, \underbrace{3.9, 4.2, 4.5, 4.8, 5.2, 5.5}_{middle}, \underbrace{5.8, 6.1, 6.4, 6.7}_{high}\}
\end{equation}
We then add $1\%$ Gaussian noise to the data. The initial model for all the inversions
is given in Fig. \ref{fig:seg_initial_model}, and is initially used as $\bfm_{ref}$ in the smoothing regularization. We apply five frequency continuation sweeps
- the first three using the low frequencies with the regularization \eqref{eq:high_order_regularization}, followed by two sweeps using low and middle frequencies with the regularization \eqref{eq:diff_regularization} to obtain a sharp model. For the first three sweeps we used 10 GN iterations per outer iteration, and for the last two we used 15 GN iterations. To finalize the inversion, we applied
up to 15 additional GN iterations involving the four highest frequencies, in order to further sharpen the reconstructed model. For all the sweeps we used 5 Conjugate Gradient (CG) iterations in each formulations.

\subsection{FWI with simultaneous sources for the SEG/EAGE model}
\begin{figure}[H]
    \centering
    \begin{subfigure}{.45\textwidth}
        \includegraphics[width=\textwidth]{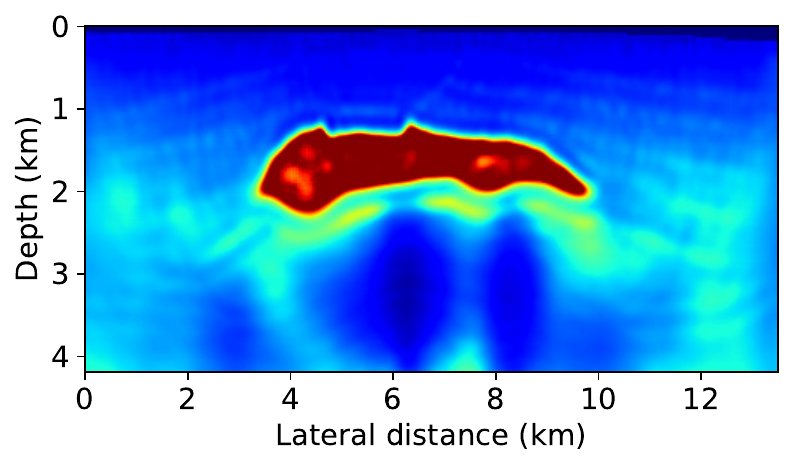}
		\caption{FWI estimated SEG/EAGE model}
        \label{fig:seg_final_model}
	\end{subfigure}
    \hspace{10pt}
    \begin{subfigure}{.45\textwidth}
        \includegraphics[width=\textwidth]{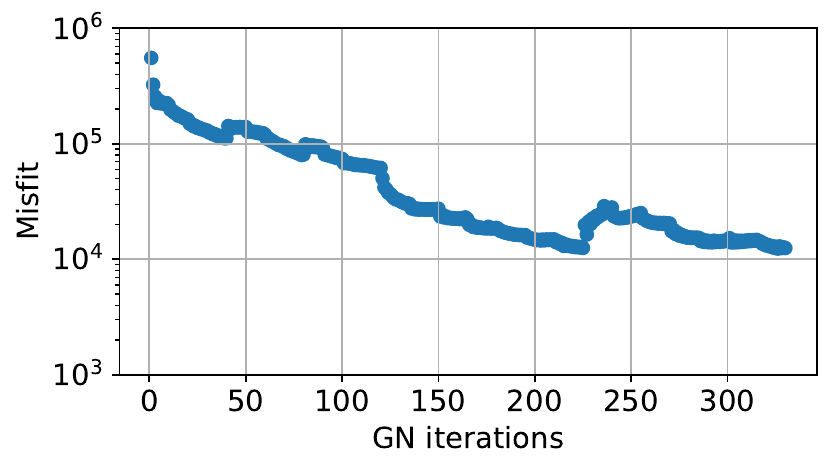}
		\caption{Misfit history}
        \label{fig:seg_misfit}
	\end{subfigure}      
\caption{\footnotesize\textit{SEG/EAGE FWI result and the corresponding total misfit history}}
\end{figure}

For the SEG/EAGE salt model experiment, we use simultaneous
sources. We set the size of $\textbf{X}$ in \eqref{eq:trace_estimation_def}
to be $n_{s} \times p$, with $p = 16$. We see that the reconstructed model,
shown in Fig. \ref{fig:seg_final_model}, estimates the ground truth salt model, \ref{fig:seg_true_model}, with high accuracy. This similarity is also evident in the total misfit history in Fig. \ref{fig:seg_misfit}.

\subsection{SEG/EAGE computational cost and solve time}
\begin{figure}[H]
\centering
\includegraphics[width=0.9\textwidth,keepaspectratio]{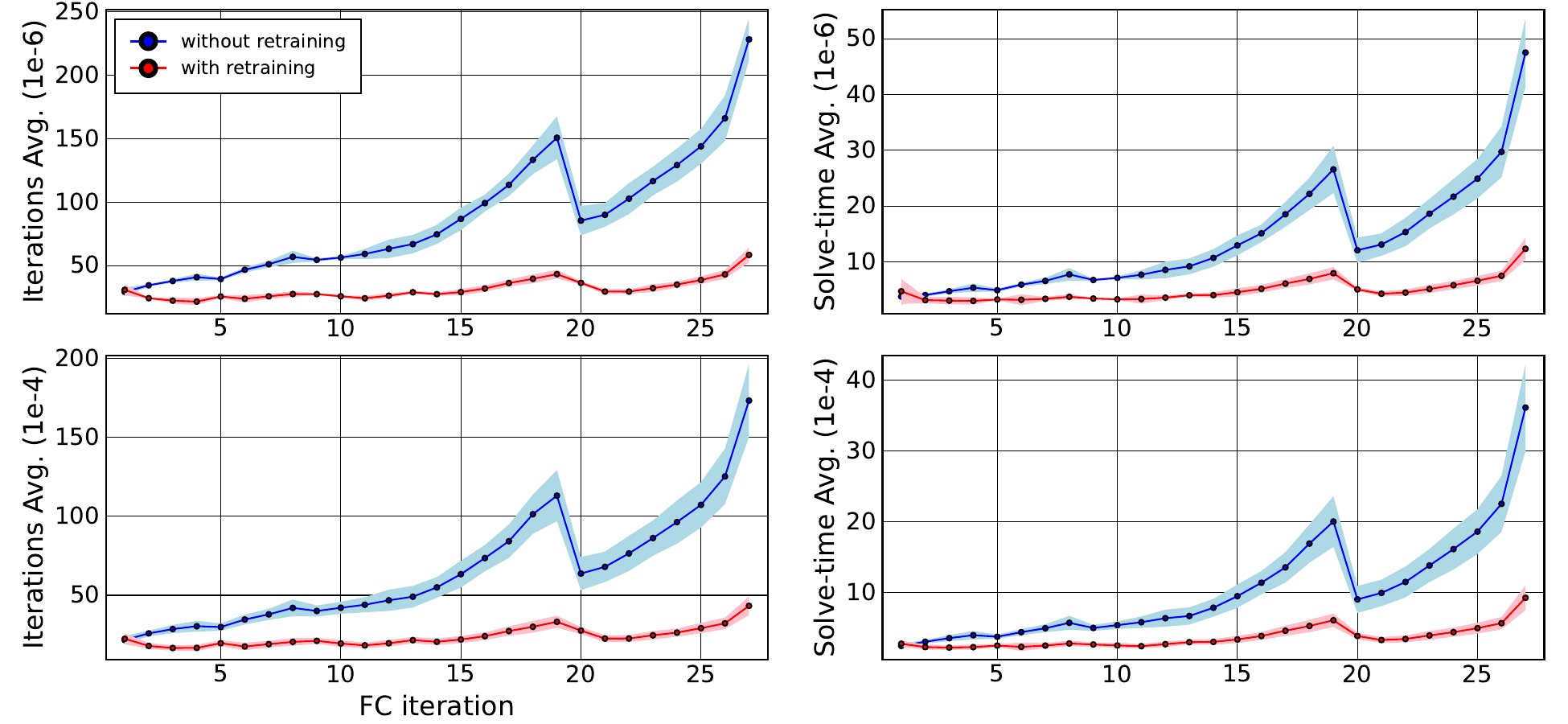}
\caption{\footnotesize\textit{SEG/EAGE salt model FWI average comparison. In the left column, we compare the average iteration count with and without training our model. The right column shows a comparison of the average solve time. Both comparisons were conducted for the different tolerance values.}}
\label{fig:avg_comparison}
\end{figure}

To compare the computational cost and the solve time of the FWI problem with and without our proposed retraining procedure we compare the iteration count and solve time of each optimization problem in the frequency continuation algorithm \ref{alg:frequency_continuation}. As described, we perform multiple frequency continuation cycles where in each cycle, we iterate over the available frequencies in a sliding window fashion and use those frequencies to solve an optimization problem. In Fig. \ref{fig:avg_comparison}, we compare the average iteration count and average solve-time of each window in the inversion process. We plot the average cost, with the corresponding standard deviation of each window, for the different tolerance thresholds, $10^{-6}$ and $10^{-4}$, used throughout the solution process for the forward model computation and sensitivities, respectively. The drop depicted in the graphs is due to the transition between the last window in the fourth cycle, containing the highest middle part frequencies, and the first window of the fifth cycle, containing the four lowest frequencies (low part). We can clearly see that by applying our light-weight retraining procedure, we obtain a computational and time efficient FWI solution process. Moreover, we notice that by applying the retraining procedure, our network has less variance, which indicates a unifying process of adapting the network to the different grid resolutions.

\begin{figure}[H]
\centering
\includegraphics[width=0.9\textwidth,keepaspectratio]{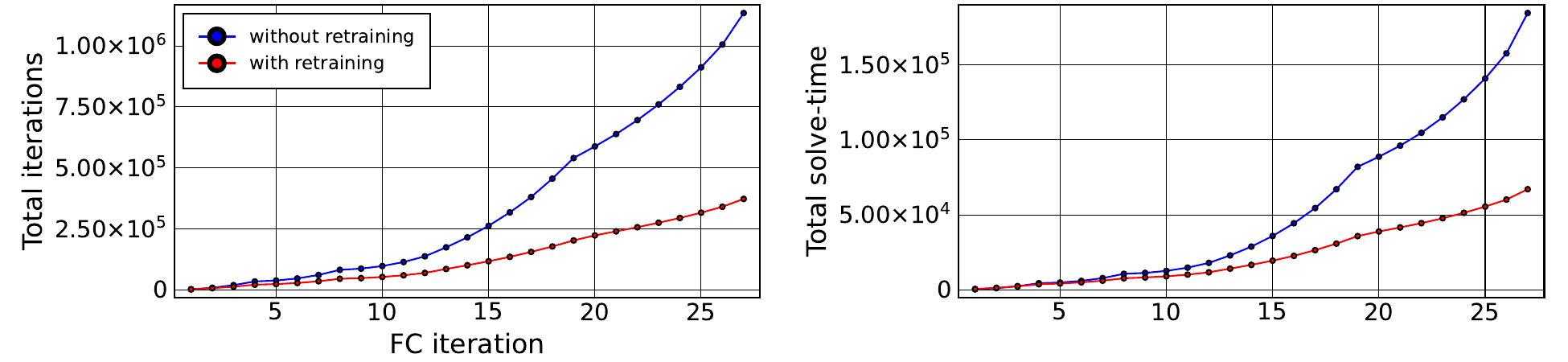}
\caption{\footnotesize\textit{SEG/EAGE salt model FWI total iteration count and solve-time comparison.}}
\label{fig:seg_total_comparison}
\end{figure}

In addition, we compare the \textbf{total} iteration count and solve-time needed for the \textbf{entire} FWI optimization process in Fig. \ref{fig:seg_total_comparison}. The graphs plot the accumulated iteration count and solve-time of the FWI process, including the time of the retraining procedure. By applying the retraining procedure we obtained the solution of the FWI process in $3.73\cdot 10^{5}$ iterations compared to $1.13\cdot 10^{6}$ without retraining, and, in a total solve-time of $6.708\cdot 10^{4}$ seconds compared to $1.846\cdot 10^{5}$. That is, we obtained an improvement of $7.57\cdot 10^{5}$ iterations and $1.2419\cdot 10^{5}$ seconds, which amounts to a $67\%$ and $63\%$ speed-up for iteration count and solve-time, respectively.


\subsection{Marmousi model setup and results} 
\begin{figure}[H]
    \centering
    \begin{subfigure}{.48\textwidth}
        \includegraphics[width=\textwidth]{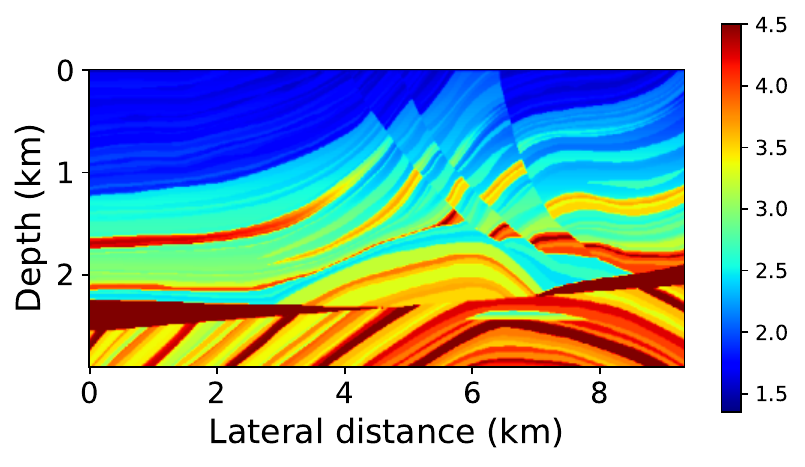}
		\caption{True velocity model}
        \label{fig:Marmousi_true_model}
	\end{subfigure}
    \begin{subfigure}{.48\textwidth}
        \includegraphics[width=\textwidth]{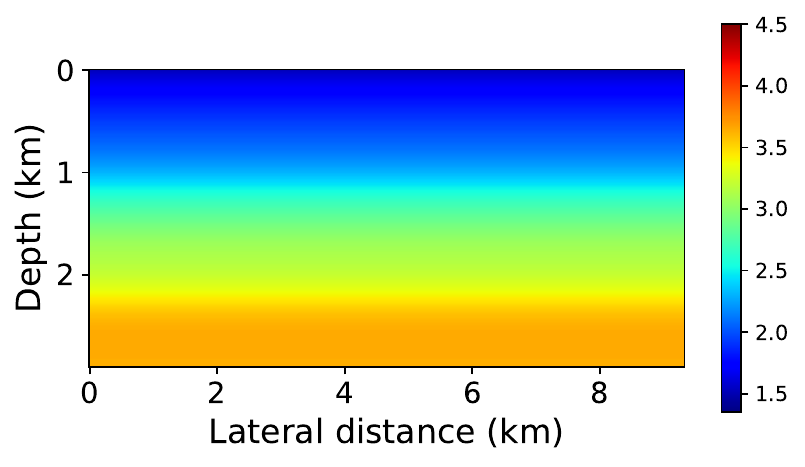}
		\caption{Reference initial model}
        \label{fig:Marmousi_initial_model}
	\end{subfigure}      
\caption{\footnotesize\textit{The 2D Marmousi velocity model and initial guess}}
\end{figure}

Our next set of experiments is conducted using
the Marmousi model presented in Fig. \ref{fig:Marmousi_true_model}. The model is described
by a $544 \times 272$ grid, representing an area of size $9.334km \times 2.904km$. The data
is generated by first solving \eqref{eq:helmholtz} for frequencies $\omega_{j} = 2\pi f_{j}$, where
\begin{equation}  
\{f_{j}\} = \{\underbrace{2.9, 3.3, 3.7, 4.2}_{low}, \underbrace{4.6, 5.0, 5.4, 5.8, 6.2, 6.7, 7.1}_{middle}, \underbrace{7.5, 7.9, 8.3, 8.7}_{high}\}.
\end{equation}
We then add $1\%$ Gaussian noise to the data. The initial model for all the inversions
is given in Fig. \ref{fig:Marmousi_initial_model}, and is initially used as $\bfm_{ref}$ in the smoothing regularization. We apply three frequency continuation sweeps
- the first sweep using the first four frequencies, followed by two sweeps using all frequencies. To finalize the inversion, we applied
up to 30 additional GN iterations involving the four highest frequencies, in order to further sharpen the reconstructed model. For all the sweeps we used 10 GN iterations and 5 Conjugate Gradient (CG) iterations with the diffusion regularization term \eqref{eq:diff_regularization}.

\subsection{FWI with simultaneous sources for the Marmousi model}
\begin{figure}[H]
    \centering
    \begin{subfigure}{.42\textwidth}
        \includegraphics[width=\textwidth]{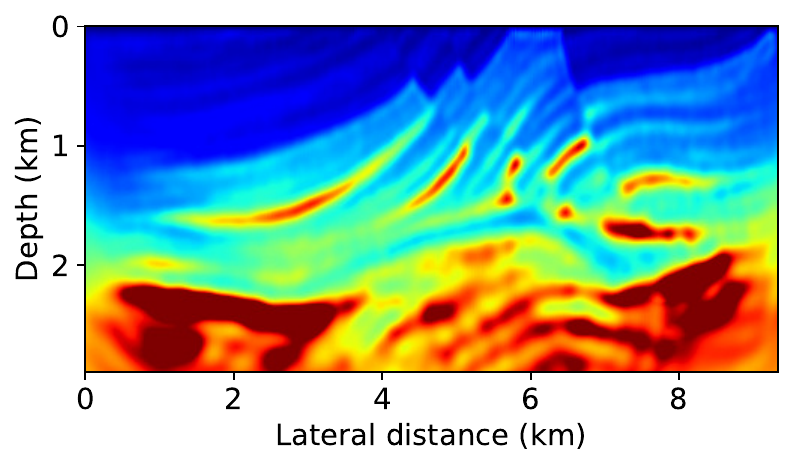}
		\caption{FWI estimated Marmousi model}
        \label{fig:Marmousi_final_model}
	\end{subfigure}
    \hspace{10pt}\begin{subfigure}{.42\textwidth}
        \includegraphics[width=\textwidth]{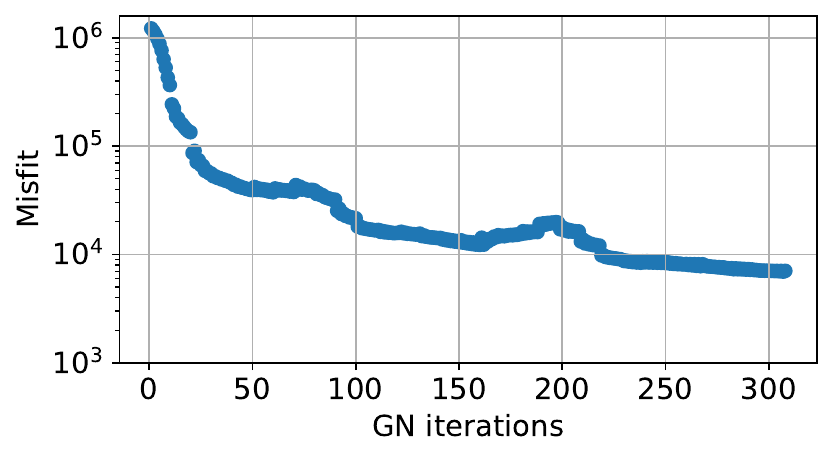}
		\caption{Misfit history}
        \label{fig:Marmousi_misfit}
	\end{subfigure}      
\caption{\footnotesize\textit{Marmousi FWI result and the corresponding misfit history}}
\end{figure}

For the Marmousi model experiment, we use simultaneous
sources and we follow the same parameters as in the SEG/EAGE model experiments. Specifically, we set $p=16$ for the simultaneous sources formulation. As before, we see that the reconstructed model,
shown in Fig. \ref{fig:Marmousi_final_model}, estimates the ground truth model, \ref{fig:Marmousi_true_model}, with high accuracy. This similarity is also evident in the misfit history in Fig. \ref{fig:Marmousi_misfit}.

\subsection{Marmousi Computational Cost and Solve Time}
\begin{figure}[H]
\centering
\includegraphics[width=0.9\textwidth,keepaspectratio]{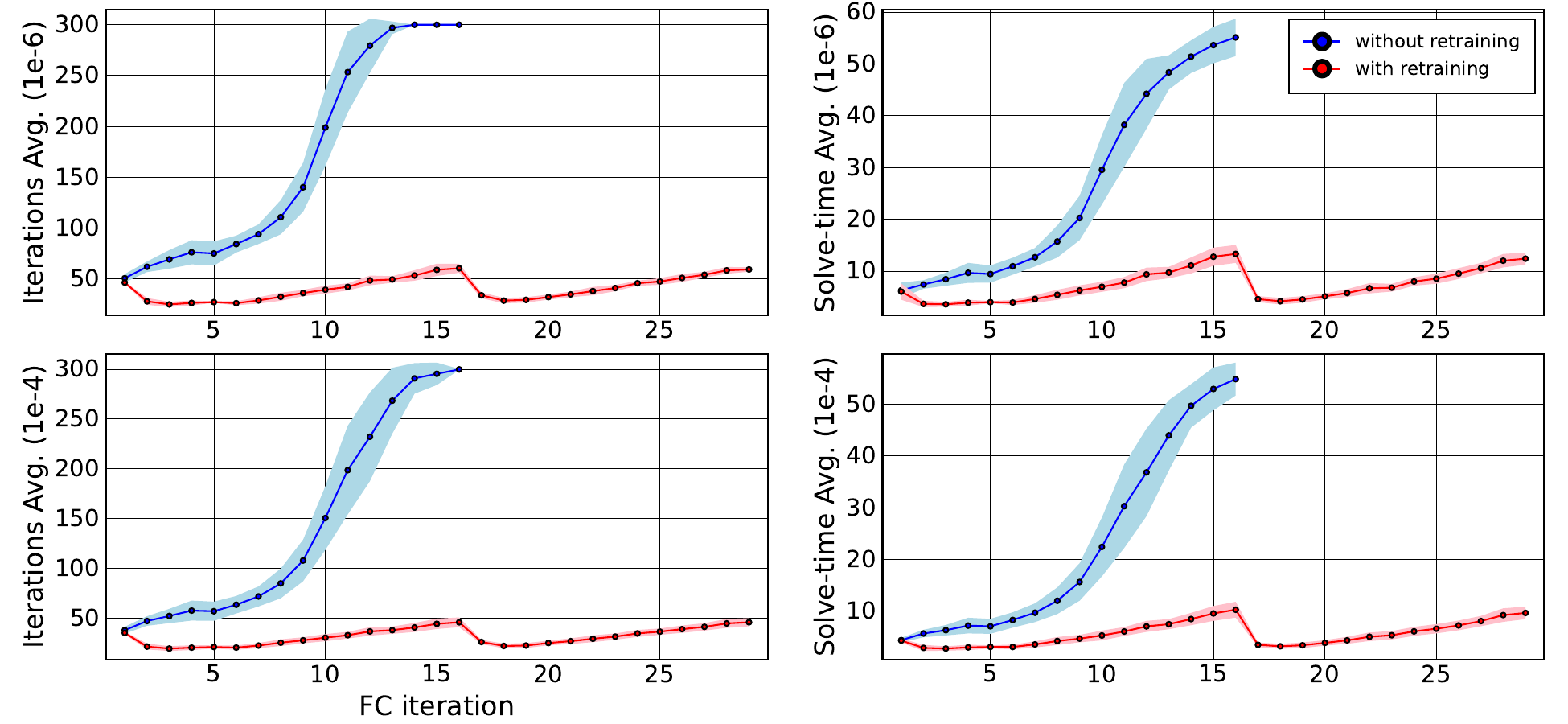}
\caption{\footnotesize\textit{Marmousi model FWI test. In the left column we compare the average iteration count with and without training our model. The right column shows a comparison of the average solve time. Both comparisons were conducted for the different tolerance values. The experiments without retraining were stopped due to the effective stagnation of the solver.}}
\label{fig:Marmousi_avg_comparison}
\end{figure}

\begin{figure}[H]
\centering
\includegraphics[width=0.9\textwidth,keepaspectratio]{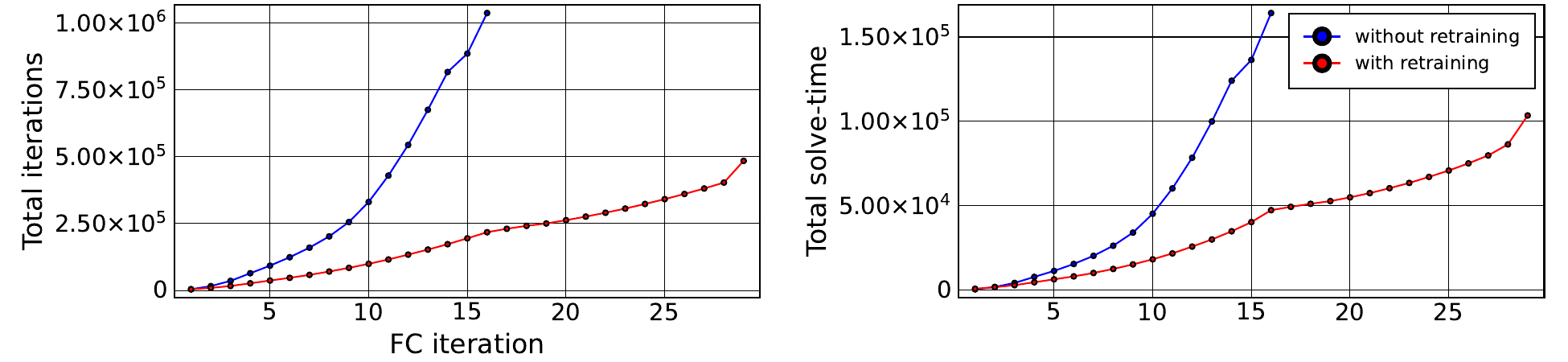}

\caption{\footnotesize\textit{Marmousi model FWI total iteration count and solve-time comparison. The experiments without retraining were stopped due to the effective stagnation of the solver.}}
\label{fig:Marmousi_total_comparison}
\end{figure}

Similarly to previous experiments, we show the comparison of solving the FWI problem with and without our retraining procedure for the Marmousi model. The Marmousi model contains high contrast areas and is highly heterogeneous. As depicted in Fig. \ref{fig:Marmousi_avg_comparison} and Fig. \ref{fig:Marmousi_total_comparison}, without adjusting the weights of the Encoder-Solver the solver reaches stagnation at an early stage of the inversion process. In our experiments we have limited the iterative solver of the Helmholtz equation to $300$ iterations, which is computationally expensive and is considered unfeasible for large scale data. However, posing a lower iteration limit to the solver will lead to a non-plausible reconstruction due to the Encoder-Solver failing to converge to the required threshold accuracy. On the other hand, when incorporating our light-weight retraining procedure, not only the Encoder-Solver does not stagnate, it also obtains low iteration counts and significantly low solve time. We see that even for highly heterogeneous medium the solver is able to cope with the gradual changes and obtain high accuracy with low iteration count. Moreover, similarly to the SEG/EAGE experiments, we notice that a unifying process of the network's performance is obtained here as well. Namely, the variance of the iteration count and solve-time remains significantly low, indicating the ability of the retraining procedure to adapt the network to different grid resolutions for the case of highly heterogeneous slowness models such as the Marmousi model.

\section{Conclusion}
In this work, we considered the integration of DL methods in the inversion process of FWI under the physics-guided approach. Namely, we integrated the Encoder-Solver model, presented in \cite{azulay2022multigrid}, in the data simulation throughout the inversion process. Our solver is applied to both the forward problem and its adjoint through manipulating the right-hand-sides. We also presented a lightweight data-efficient retraining procedure to the Encoder-Solver model. The retraining procedure allows the efficient integration of the Encoder-Solver model in the inversion process of the FWI problem. Our proposed retraining procedure demonstrated significant improvements to the Encoder-Solver performance, especially when solving for highly contrasted mediums such as the Marmousi model. Together with source-encoding, this yields the solution of physics-guided FWI at a more reasonable cost than traditional methods. 

%Moreover, we have shown that our network can solve the adjoint system by performing a simple modification to the right-hand-side vectors. Such property simplifies the training and retraining procedures and further introduces efficiency to the inversion process.

%\end{document}

\bibliographystyle{plain}
\bibliography{main.bib}

\end{document}